\documentclass{article}
\usepackage{ijcai20}

\usepackage{times}
\usepackage{soul}
\usepackage{url}
\usepackage[hidelinks]{hyperref}
\usepackage[utf8]{inputenc}
\usepackage[small]{caption}
\usepackage{graphicx}
\usepackage{amsmath}
\usepackage{amsthm}
\usepackage{booktabs}
\usepackage{algorithm}
\usepackage{bbm}
\usepackage[linewidth=1pt]{mdframed}
\usepackage{tikz-dependency}
\usepackage{amsmath, graphicx, amsfonts, amssymb}
\usepackage{algorithmicx}
\usepackage{algorithm}
\usepackage{booktabs}
\usepackage[noend]{algpseudocode}
\usepackage[percent]{overpic}
\usepackage{subcaption}
\usepackage{multirow}
\usepackage[page]{appendix}

\usepackage{listings}
\NewEnviron{elaboration}{
\par
\begin{tikzpicture}
\node[rectangle,minimum width=\textwidth] (m) {\begin{minipage}{.98\textwidth}\BODY\end{minipage}};
\draw[dashed] (m.south west) rectangle (m.north east);
\end{tikzpicture}
}
\usepackage{xcolor}

\title{Bringing Stories Alive: Generating Interactive Fiction Worlds}

\author{
Prithviraj Ammanabrolu\footnote{Denotes equal contribution.}
\and
Wesley Cheung$^*$\and
Dan Tu\and \\
William Broniec \And
Mark O. Riedl
\affiliations
School of Interactive Computing\\
Georgia Institute of Technology
\emails
\{raj.ammanabrolu, wcheung8, dan.tu wbroniec3, riedl\}@gatech.edu,}

\newcommand{\citet}[1]{\citeauthor{#1}~\shortcite{#1}}
\begin{document}

\maketitle
\begin{abstract}
World building forms the foundation of any task that requires narrative intelligence.
In this work, we focus on procedurally generating interactive fiction worlds---text-based worlds that players ``see'' and ``talk to'' using natural language.
Generating these worlds requires referencing everyday and thematic commonsense priors in addition to being semantically consistent, interesting, and coherent throughout.
Using existing story plots as inspiration, we present a method that first extracts a partial knowledge graph encoding basic information regarding world structure such as locations and objects.
This knowledge graph is then automatically completed utilizing thematic knowledge and used to guide a neural language generation model that fleshes out the rest of the world.
We perform human participant-based evaluations, testing our neural model's ability to extract and fill-in a knowledge graph and to generate language conditioned on it against rule-based and human-made baselines.
Our code is available at \url{https://github.com/rajammanabrolu/WorldGeneration}.
\end{abstract}

\section{Introduction}

\begin{figure}[htp]
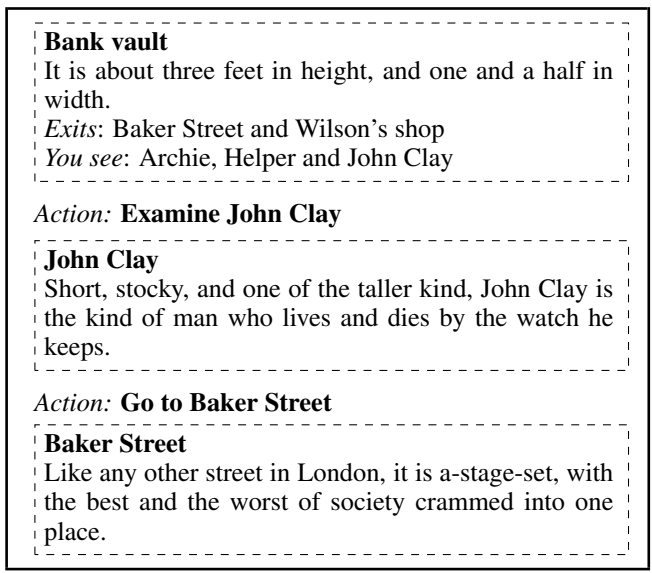

\begin{mdframed}
\begin{elaboration}
  \parbox{.99\textwidth}{
\textbf{Bank vault}\\
It is about three feet in height, and one and a half in width.\\
\emph{Exits}: Baker Street and Wilson's shop\\
\emph{You see}: Archie, Helper and John Clay
}
\end{elaboration}
\begin{flushleft}
\emph{Action:} \textbf{Examine John Clay}
\end{flushleft}
\begin{elaboration}
  \noindent\parbox{.99\textwidth}{
\textbf{John Clay}\\
Short, stocky, and one of the taller kind, John Clay is the kind of man who lives and dies by the watch he keeps.
}
\end{elaboration}
\begin{flushleft}
\emph{Action:} \textbf{Go to Baker Street}
\end{flushleft}
\begin{elaboration}
  \noindent\parbox{.99\textwidth}{
\textbf{Baker Street}\\
Like any other street in London, it is a-stage-set, with the best and the worst of society crammed into one place.
}
\end{elaboration}
\end{mdframed}
\caption{Example player interaction in the deep neural generated mystery setting.}
\label{fig:descriptions}
\end{figure}

{\em Interactive fictions}---also called text-adventure games or text-based games---are games in which a player interacts with a virtual world purely through textual natural language---receiving descriptions of what they ``see'' and writing out how they want to act, an example can be seen in Figure~\ref{fig:descriptions}.
Interactive fiction games are often structured as puzzles, or quests, set within the confines of given game world.
%
Interactive fictions have been adopted as a test-bed for real-time game playing agents~\cite{Narasimhan2015,cote2018textworld,hausknecht}. 
Unlike other, graphical games, interactive fictions test agents' abilities to infer the state of the world through communication and to indirectly affect change in the world through language.  
Interactive fictions are typically modeled after real or fantasy worlds; commonsense knowledge is an important factor in successfully playing interactive fictions~\cite{ammanabrolutransfer,jonmay}.

In this paper we explore a different challenge for artificial intelligence: automatically generating text-based virtual worlds for interactive fictions. 
A core component of many narrative-based tasks---everything from storytelling to game generation---is world building.
The world of a story or game defines the boundaries of where the narrative is allowed and what the player is allowed to do.
There are four core challenges to world generation: 
(1)~commonsense knowledge: 
the world must reference priors that the player possesses so that players can make sense of the world and build expectations on how to interact with it.
This is especially true in interactive fictions where the world is presented textually because many details of the world necessarily be left out (e.g., the pot is on a stove; kitchens are found in houses) that might otherwise be literal in a graphical virtual world.
(2)~Thematic knowledge: 
interactive fictions usually involve a theme or genre that comes with its own expectations. 
For example, light speed travel is plausible in sci-fi worlds but not realistic in the real world.
(3)~Coherence: the world must not appear to be an random assortment of locations.
(3)~Natural language: The descriptions of the rooms as well as the permissible actions must text, implying that the system has  natural language generation capability.

Because worlds are conveyed entirely through natural language, the potential output space for possible generated worlds is combinatorially large. 
To constrain this space and to make it possible to evaluate generated world, we present an approach which makes use of existing stories, building on the worlds presented in them but leaving enough room for the worlds to be unique.
Specifically, we take a story such as Sherlock Holmes or Rapunzel---a linear reading experience---and extract the description of the world the story is set in to make an interactive world the player can explore.

Our method first extracts a partial, potentially disconnected knowledge graph from the story, encoding information regarding locations, characters, and objects in the form of $\langle entity,relation,entity\rangle$ triples.
Relations between these types of entities as well as their properties are captured in this knowledge graph.
However, stories often do not explicitly contain all the information required to fully fill out such a graph.
A story may mention that there is a sword stuck in a stone but not what you can do with the sword or where it is in relation to everything else.
Our method fills in missing relation and affordance information using thematic knowledge gained from training on stories in a similar genre.
This knowledge graph is then used to guide the text description generation process for the various locations, characters, and objects.
The game is then assembled on the basis of the knowledge graph and the corresponding generated descriptions.


We have two major contributions.
(1) A neural model and a rules-based baseline for each of the tasks described above.
The phases are that of graph extraction and completion followed by description generation and game formulation.
Each of these phases are relatively distinct and utilize their own models.
(2) A human subject study for comparing the neural model and variations on it to the rules-based and human-made approaches.
We perform two separate human subject studies---one for the first phase of knowledge graph construction and another for the overall game creation process---testing specifically for coherence, interestingness, and the ability to maintain a theme or genre.


\section{Related Work}



There has been a slew of recent work in developing agents that can play text games~\cite{Narasimhan2015,Haroush2018,cote2018textworld,jericho}.
\citeauthor{ammanabrolu}~\shortcite{ammanabrolutransfer,ammanabrolu,ammanabrolu2020graph} in particular use knowledge graphs as state representations for game-playing agents.
\citet{yuan2019qait} propose QAit, a set of question answering tasks framed as text-based or interactive fiction games.
QAit focuses on helping agents learn procedural knowledge through interaction with a dynamic environment.
These works all focus on agents that learn to play a given set of interactive fiction games as opposed to generating them.


Scheherazade~\cite{Li2012} is a system that learns a plot graph based on stories written by crowd sourcing the task of writing short stories.
The learned plot graph contains details relevant to ensure story coherence.
It includes: plot events, temporal precedence, and mutual exclusion relations.
Scheherazade-IF~\cite{guzdial2015crowdsourcing} extends the system to generate choose-your-own-adventure style interactive fictions in which the player chooses from prescribed options.
\citet{womakfreemanifgen} explore a method of creating interactive narratives revolving around locations, wherein sentences are mapped to a real-world GPS location from a corpus of sentences belonging to a certain genre.
Narratives are made by chaining together sentences selected based on the player's current real-world location.
In contrast to these models, our method generates a parser-based interactive fiction in which the player types in a textual command, allowing for greater expressiveness.

\citet{ammanabrolu2019toward} define the problem of procedural content generation in interactive fiction games in terms of the twin considerations of world and quest generation and focus on the latter.
They present a system in which {\em quest} content is first generated by learning from a corpus and then grounded into a given interactive fiction world.
The work is this paper focuses on the {\em world} generation problem glossed in the prior work. 
Thus these two systems can be seen as complimentary.

Light~\cite{Urbanek2019LearningTS} is a crowdsourced dataset of grounded text-adventure game dialogues.
It contains information regarding locations, characters, and objects set in a fantasy world.
The authors demonstrate that the supervised training of transformer-based models lets us contextually relevant dialog, actions, and emotes.
Most in line with the spirit of this paper, \citet{fan2019generating} leverage Light to generate worlds for text-based games.
They train a neural network based model using Light to compositionally arrange locations, characters, and objects into an interactive world.
Their model is tested using a human subject study against other machine learning based algorithms with respect to the cohesiveness and diversity of generated worlds.
Our work, in contrast, focuses on extracting the information necessary for building interactive worlds from existing story plots.


\section{World Generation}
\label{sec:worldgen}

World generation happens in two phases.
In the first phase, a partial knowledge graph is extracted from a story plot and then filled in using thematic commonsense knowledge.
In the second phase, the graph is used as the skeleton to generate a full interactive fiction game---generating textual descriptions or ``flavortext'' for rooms and embedded objects.
We present a novel neural approach in addition to a rule guided baseline for each of these phases in this section.

\subsection{Knowledge Graph Construction}
\begin{figure}
    \centering
    \includegraphics[width=0.9\linewidth]{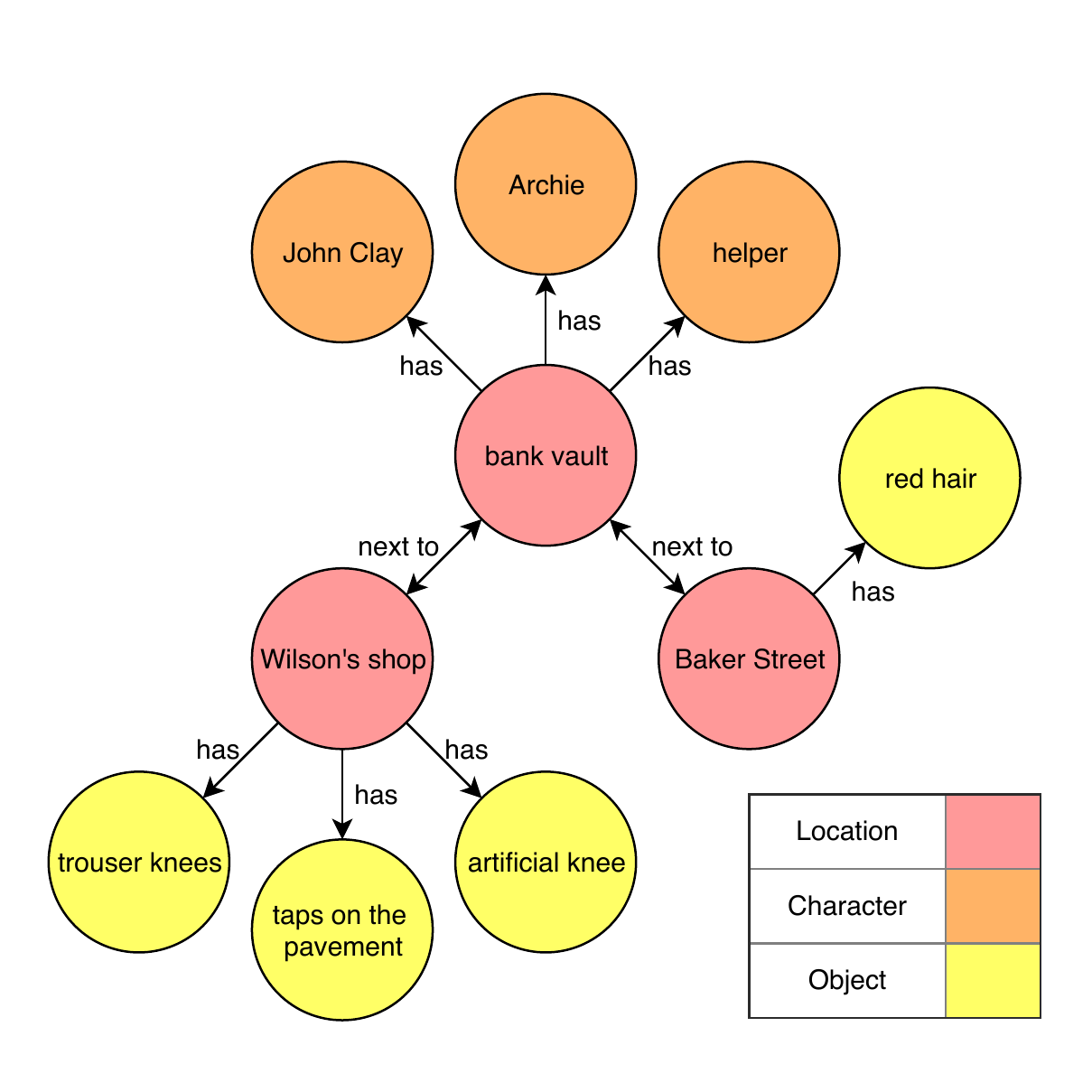}
    
    \vspace{-10pt}
    \caption{Example knowledge graph constructed by AskBERT.}
    \label{fig:kg}
\end{figure}

The first phase is to extract a knowledge graph from the story that depicts locations, characters, objects, and the relations between these entities.
We present two techniques.
The first uses neural question-answering technique to extract relations from a story text.
The second, provided as a baseline, uses  OpenIE5\footnote{\url{https://github.com/dair-iitd/OpenIE-standalone}}, a commonly used rule-based information extraction technique. 
For the sake of simplicity, we considered primarily the location-location and location-character/object relations, represented by the ``next to'' and ``has'' edges respectively in Figure \ref{fig:kg}.

\subsubsection{Neural Graph Construction}

\begin{figure}
\centering
\includegraphics[width=\linewidth]{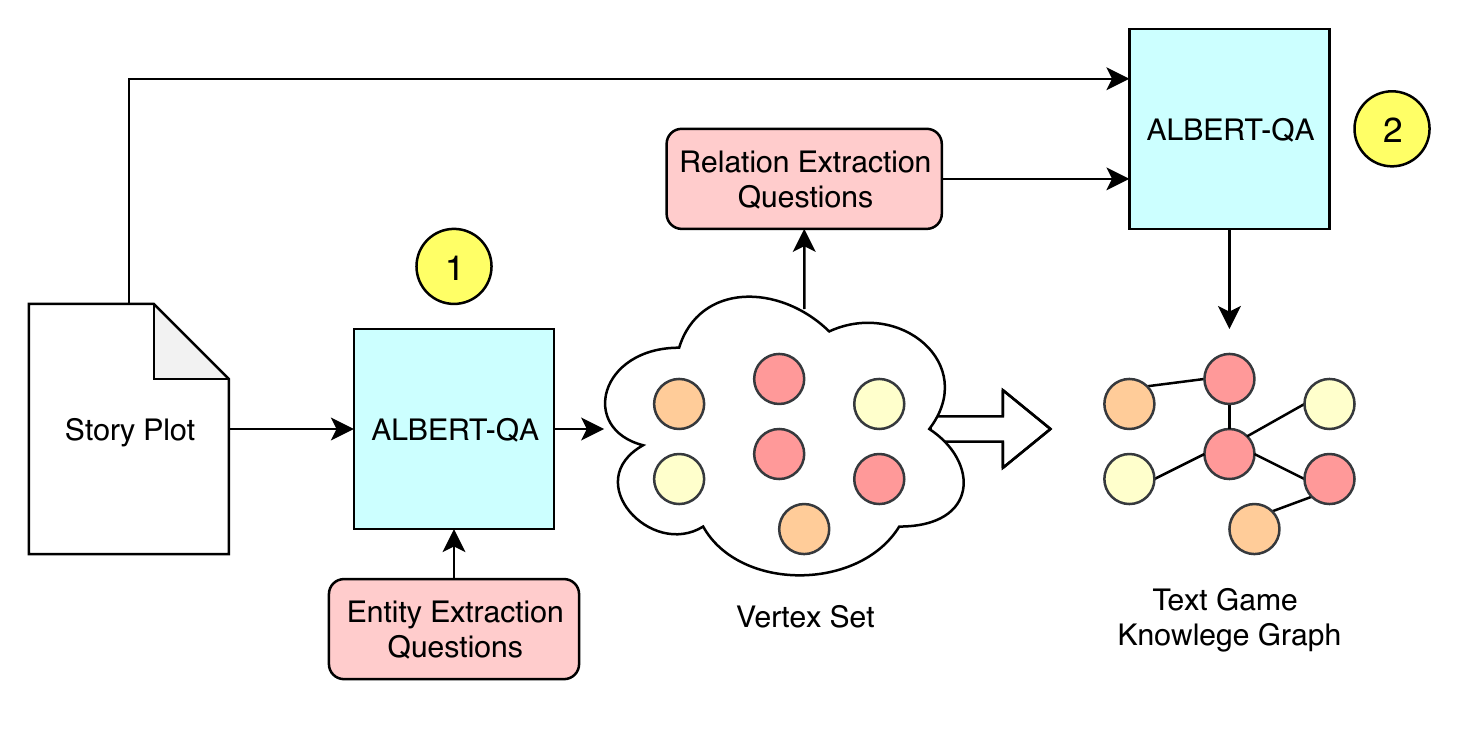}
\vspace{-10pt}
\caption{Overall AskBERT pipeline for graph construction. }%
\label{fig:pipeline}
\end{figure}

While many neural models already exist that perform similar tasks such as named entity extraction and part of speech tagging, they often come at the cost of large amounts of specialized labeled data suited for that task. 
We instead propose a new method that leverages models trained for context-grounded question-answering tasks to do entity extraction with no task dependent data or fine-tuning necessary.
Our method, dubbed {\em AskBERT}, leverages the Question-Answering (QA) model ALBERT~\cite{lan2019}.
AskBERT consists of two main steps as shown in Figure~\ref{fig:pipeline}: vertex extraction and graph construction.

The first step is to extract the set of entities---graph vertices---from the story.
We are looking to extract information specifically regarding characters, locations, and objects.
This is done by using asking the QA model questions such as ``Who is a character in the story?''.
\citet{ribeiro2019red} have shown that the phrasing of questions given to a QA model is important and this forms the basis of how we formulate our questions---questions are asked so that they are more likely to return a single answer, e.g. asking ``Where is a location in the story?'' as opposed to ``Where are the locations in the story?''. In particular, we notice that pronoun choice can be crucial; ``Where is a location in the story?'' yielded more consistent extraction than ``What is a location in the story?''.
ALBERT QA is trained to also output a special \textless$no$-$answer$\textgreater \ token when it cannot find an answer to the question within the story.
Our method makes use of this by iteratively asking QA model a question and masking out the most likely answer outputted on the previous step.
This process continues until the \textless$no$-$answer$\textgreater \ token becomes the most likely answer.

The next step is graph construction.
Typical interactive fiction worlds are usually structured as trees, i.e. no cycles except between locations.
Using this fact, we use an approach that builds a graph from the vertex set by one relation---or edge---at a time.
Once again using the entire story plot as context, we query the ALBERT-QA model picking a random starting location $x$ from the set of vertices previously extracted.
and asking the questions ``What location can I visit from $x$?'' and ``Who/What is in $x$?''.
The methodology for phrasing these questions follows that described for the vertex extraction.
The answer given by the QA model is matched to the vertex set by picking the vertex $u$ that contains the best word-token overlap with the answer.
Relations between vertices are added by computing a relation probability on the basis of the output probabilities of the answer given by the QA model.
The probability that vertices $x,u$ are related:
\begin{equation}
    P(x,u) = \frac{p(x,u) + p(u,x)}{2}
\end{equation}
where
\begin{equation}
p(x,u) = \sum_{o \in \text{QA outputs}} p(o)\mathbbm{1}\{u = \underset{v}{\mathrm{argmax}}(v\cap o)\}
\end{equation}
is the sum of the individual token probabilities of all the overlapping tokens in the answer from the QA model and $u$. 

\subsubsection{Rule-Based Graph Construction}

We compared our proposed AskBERT method with a non-neural, rule-based approach. This approach is based on the information extracted by OpenIE5, followed by some post-processing such as named-entity recognition and part-of-speech tagging. 
OpenIE5 combines several cutting-edge ideas from several existing papers~\cite{saha2018open,pal2016demonyms,christensen2011analysis} to create a powerful information extraction tools. 
For a given sentence, OpenIE5 generates multiple triples in the format of $\langle entity, relation, entity\rangle$ as concise representations of the sentence, each with a confidence score. 
These triples are also occasionally annotated with location information indicating that a triple happened in a location.




As in the neural AskBERT model, we attempt to extract information regarding locations, characters, and objects.
The entire story plot is passed into the OpenIE5 and we receive a set of triples.
The location annotations on the triples are used to create a set of locations.
We mark which sentences in the story contain these locations.
POS tagging based on marking noun-phrases is then used in conjunction with NER to further filter the set of triples---identifying the set of characters and objects in the story.

\begin{figure*}[t]
\centering
\includegraphics[width=0.8\linewidth]{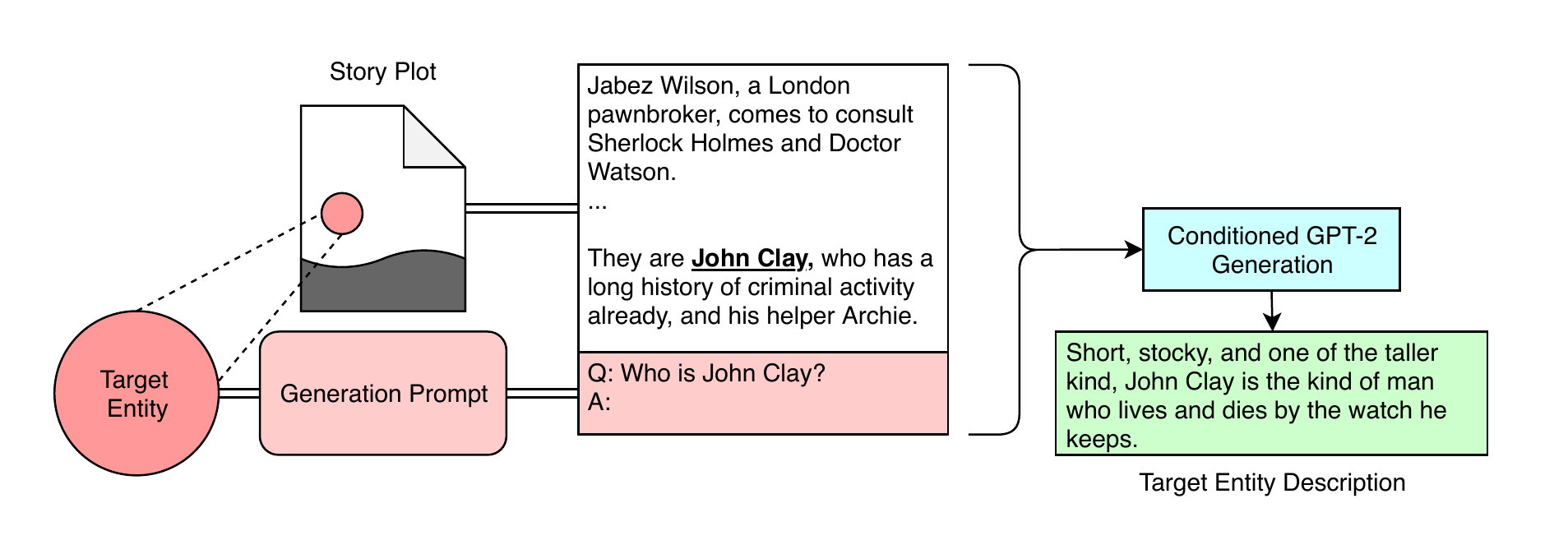}
\vspace{-10pt}
\caption{Overview for neural description generation.}%
\label{fig:generation}
\end{figure*}

The graph is constructed by linking the set of triples on the basis of the location they belong to.
While some sentences contain very explicit location information for OpenIE5 to mark it out in the triples, most of them do not.
We therefore make the assumption that the location remains the same for all triples extracted in between sentences where locations are explicitly mentioned.
For example, if there exists $location A$ in the 1st sentence and $location B$ in the 5th sentence of the story, all the events described in sentences 1-4 are considered to take place in $location A$. 
The entities mentioned in these events are connected to $location A$ in the graph.

\subsection{Description Generation}


The second phase involves using the constructed knowledge graph to generate textual descriptions of the entities we have extracted, also known as flavortext.
This involves generating descriptions of what a player ``sees'' when they enter a location and short blurbs for each object and character.
These descriptions need to not only be faithful to the information present in the knowledge graph and the overall story plot but to also contain flavor and be interesting for the player.

\subsubsection{Neural Description Generation} 
Here, we approach the problem of description generation by taking inspiration from conditional transformer-based generation methods~\cite{shirish2019}.
Our approach is outlined in Figure~\ref{fig:generation} and an example description shown in Figure~\ref{fig:descriptions}.
For any given entity in the story, we first locate it in the story plot and then construct a prompt which consists of the entire story up to and including the sentence when the entity is first mentioned in the story followed by a question asking to describe that entity.
With respect to prompts, we found that more direct methods such as question-answering were more consistent than open-ended sentence completion.  
For example, ``Q: Who is the prince? A:'' often produced descriptions that were more faithful to the information already present about the prince in the story than  ``You see the prince. He is/looks''. 
For our transformer-based generation, we use a pre-trained 355M GPT-2 model \cite{radford2019language} finetuned on a corpus of plot summaries collected from Wikipedia.
The plots used for finetuning are tailored specific to the genre of the story in order to provide more relevant generation for the target genre.
Additional details regarding the datasets used are provided in Section~\ref{sec:expts}.
This method strikes a balance between knowledge graph verbalization techniques which often lack ``flavor'' and open ended generation which struggles to maintain semantic coherence.

\subsubsection{Rules-Based Description Generation}
In the rule-based approach, we utilized the templates from the built-in text game generator of TextWorld~\cite{cote2018textworld} to generate the description for our graphs. 
TextWorld is an open-source library that provides a way to generate text-game learning environments for training reinforcement learning agents using pre-built grammars. 

Two major templates involved here are the Room Intro Templates and Container Description Templates from TextWorld, responsible for generating descriptions of locations and blurbs for objects/characters respectively.
The location and object/character information are taken from the knowledge graph constructed previously.
\begin{itemize}
    \item Example of Room Intro Templates: ``This might come as a shock to you, but you've just $\#entered\#$ a \textless $location$-$name$\textgreater''
    \item Example of Container Description Templates: ``The \textless $location$-$name$\textgreater \ $\#contains\#$ \textless $object/person$-$name$\textgreater''
\end{itemize}
Each token surrounded by $\#$ sign can be expanded using a select set of terminal tokens. 
For instance, $\#entered\#$ could be filled with any of the following phrases here: entered; walked into; fallen into; moved into; stumbled into; come into. 
Additional prefixes, suffixes and adjectives were added to increase the relative variety of descriptions. 
%
Unlike the neural methods, the rule-based approach is not able to generate detailed and flavorful descriptions of the properties of the locations/objects/characters. 
By virtue of the templates, however, it is much better at maintaining consistency with the information contained in the knowledge graph.


\section{Evaluation}
\label{sec:expts}
We conducted two sets of human participant evaluations by recruiting participants over Amazon Mechanical Turk.
The first evaluation tests the knowledge graph construction phase, in which we measure perceived coherence and genre or theme resemblance of graphs extracted by different models.
The second study compares full games---including description generation and game assembly, which can't easily be isolated from graph construction---generated by different methods. This study looks at how interesting the games were to the players in addition to overall coherence and genre resemblance.
Both studies are performed across two genres: mystery and fairy-tales.
This is done in part to test the relative effectiveness of our approach across different genres with varying thematic commonsense knowledge.
The dataset used was compiled via story summaries that were scraped from Wikipedia via a recursive crawling bot.
The bot searched pages for both for plot sections as well as links to other potential stories.
From the process, 695 fairy-tales and 536 mystery stories were compiled from two categories: novels and short stories.
We note that the mysteries did not often contain many fantasy elements, i.e. they consisted of mysteries set in our world such as Sherlock Holmes, while the fairy-tales were much more removed from reality.
Details regarding how each of the studies were conducted and the corresponding setup are presented below.

\subsection{Knowledge Graph Construction Evaluation}

We first select a subset of 10 stories randomly from each genre and then extract a knowledge graph using three different models.
Each participant is presented with the three graphs extracted from a single story in each genre and then asked to rank them on the basis of how coherent they were and how well the graphs match the genre. 
The graphs resembles the one shown in in Figure~\ref{fig:kg} and are presented to the participant sequentially.
The exact order of the graphs and genres was also randomized to mitigate any potential latent correlations.
Overall, this study had a total of 130 participants.
This ensures that, on average, graphs from every story were seen by 13 participants.

\begin{table}[t]
\scriptsize
\centering
\begin{tabular}{l|l|l|l}
Genre                       & Category   & Neural & Rules \\ \hline
\multirow{3}{*}{Mystery}    & Locations  & 7.2    & 3.5   \\
                            & Characters & 4.8    & 4.1   \\
                            & Objects    & 3.2    & 12.2  \\ \hline
\multirow{3}{*}{Fairy-tale} & Locations  & 4.0    & 1.8   \\
                            & Characters & 3.3    & 1.2   \\
                            & Objects    & 4.1    & 8.7   

\end{tabular}
\caption{\textbf{Vertex statistics:} Average vertex count by type per genre. The random model has the same vertex statistics as the neural model.}
\label{tab:ablation}

\end{table}

\begin{table}[t]
\scriptsize
\centering
\begin{tabular}{l|l|l|l|l}
\multicolumn{1}{l|}{Genre} & \multicolumn{1}{l|}{Statistic} & \multicolumn{1}{l|}{Neural} & \multicolumn{1}{l|}{Rules} & \multicolumn{1}{l}{Random} \\ \hline
\multirow{2}{*}{Mystery}    & Avg. Edges                     & 10.7                        & 22.3                       & 10.7                        \\
                            & Avg. Degree                    & 1.63 $\pm$ 1.77             & 2.15 $\pm$ 0.38            & 1.63 $\pm$ 1.63             \\ \hline
\multirow{2}{*}{Fairy-tale} & Avg. Edges                     & 16.7                        & 12                         & 16.7                        \\
                            & Avg. Degree                    & 1.73 $\pm$ 2.04             & 1.98 $\pm$ 0.29            & 1.73 $\pm$ 1.64            
\end{tabular}
\caption{\textbf{Edge and degree statistics:} Average edge count , average degree count, and degree standard deviation of the graphs per genre.}
\label{tab: edgedegree}
\end{table}

In addition to the neural AskBERT and rules-based methods, we also test a variation of the neural model which we dub to be the ``random'' approach. The method of vertex extraction remains identical to the neural method, but we instead connect the vertices randomly instead of selecting the most confident according to the QA model. We initialize the graph with a starting location entity. Then, we randomly sample from the vertex set and connect it to a randomly sampled location in the graph until every vertex has been connected.
This ablation in particular is designed to test the ability of our neural model to predict relations between entities.
It lets us observe how accurately linking related vertices effects each of the metrics that we test for.
%
%
For a fair comparison between the graphs produced by different approaches, we randomly removed some of the nodes and edges from the initial graphs so that the maximum number of locations per graph and the maximum number of objects/people per location in each story genre are the same.

The results are shown in Table~\ref{table:kgresults}.
We show the median rank of each of the models for both questions across the genres.
Ranked data is generally closely interrelated and so we perform Friedman's test between the three models to validate that the results are statistically significant.
This is presented as the $p$-value in table (asterisks indicate significance at $p<0.05$).
In cases where we make comparisons between specific pairs of models, when necessary, we additionally perform the Mann-Whitney U test to ensure that the rankings differed significantly. 

In the mystery genre, the rules-based method was often ranked first in terms of genre resemblance, followed  by the neural and random models.
This particular result was not statistically significant however, likely indicating that all the models performed approximately equally in this category.
The neural approach was deemed to be the most coherent followed by the rules and random.
For the fairy-tales, the neural model ranked higher on both of the questions asked of the participants.
In this genre, the random neural model also performed better than the rules based approach.

Tables~\ref{tab:ablation} and~\ref{tab: edgedegree} show the statistics of the constructed knowledge graphs in terms of vertices and edges.
We see that the rules-based graph construction has a lower number of locations, characters, and relations between entities but far more objects in general.
The greater number of objects is likely due to the rules-based approach being unable to correctly identify locations and characters.
The gap between the methods is less pronounced in the mystery genre as opposed to the fairy-tales, in fact the rules-based graphs have more relations than the neural ones.
The random and neural models have the same number of entities in all categories by construction but random in general has lower variance on the number of relations found.
In this case as well, the variance is lower for mystery as opposed to fairy-tales.
When taken in the context of the results in Table~\ref{table:kgresults}, it appears to indicate that leveraging thematic commonsense in the form of AskBERT for graph construction directly results in graphs that are more coherent and maintain genre more easily.
This is especially true in the case of the fairy-tales where the thematic and everyday commonsense diverge more than than in the case of the mysteries.

\begin{table}[t]
\scriptsize
\centering
\begin{tabular}{l|l|l|l|l|l}
   Genre                         &      Questions           & Neural     & Rules      & Random & p-value        \\ \hline
\multirow{2}{*}{Mystery}    & Resembles Genre & 2          & \textbf{1} & 3      & 0.35           \\
                            & Coherence     & \textbf{1} & 2          & 3      & 0.049$^*$ \\ \hline
\multirow{2}{*}{Fairy-tale} & Resembles Genre & \textbf{1} & 3          & 2      & 0.014$^*$ \\
                            & Coherence     & \textbf{1} & 3          & 2      & 0.013$^*$
\end{tabular}
\caption{Results of the knowledge graph evaluation study.}
\label{table:kgresults}
\end{table}

\subsection{Full Game Evaluation}
This participant study was designed to test the overall game formulation process encompassing both phases described in Section~\ref{sec:worldgen}.
A single story from each genre was chosen by hand from the 10 stories used for the graph evaluation process.
From the knowledge graphs for this story, we generate descriptions using the neural, rules, and random approaches described previously.
Additionally, we introduce a human-authored game for each story here to provide an additional benchmark. 
This author selected was familiar with text-adventure games in general as well as the genres of detective mystery and fairy tale. 
To ensure a fair comparison, we ensure that the maximum number of locations and maximum number of characters/objects per location matched the other methods.
After setting general format expectations, the author read the selected stories and constructed knowledge graphs in a corresponding three step process of: identifying the $n$ most important entities in the story, mapping positional relationships between entities, and then synthesizing flavor text for the entities based off of said location, the overall story plot, and background topic knowledge.

Once the knowledge graph and associated descriptions are generated for a particular story, they are then automatically turned into a fully playable text-game using the text game engine Evennia\footnote{\url{http://www.evennia.com/}}.
Evennia was chosen for its flexibility and customization, as well as a convenient web client for end user testing. 
The data structures were translated into builder commands within Evennia that constructed the various layouts, flavor text, and rules of the game world. 
Users were placed in one “room” out of the different world locations within the game they were playing, and asked to explore the game world that was available to them.
Users achieved this by moving between rooms and investigating objects. 
Each time a new room was entered or object investigated, the player's total number of explored entities would be displayed as their score.

Each participant was was asked to play the neural game and then another one from one of the three additional models within a genre.
The completion criteria for each game is collect half the total score possible in the game, i.e. explore half of all possible rooms and examine half of all possible entities.
This provided the participant with multiple possible methods of finishing a particular game.
On completion, the participant was asked to rank the two games according to overall perceived coherence, interestingness, and adherence to the genre. 
We additionally provided a required initial tutorial game which demonstrated all of these mechanics.
The order in which participants played the games was also randomized as in the graph evaluation to remove potential correlations.
We had 75 participants in total, 39 for mystery and 36 for fairy-tales.
As each player played the neural model created game and one from each of the other approaches---this gave us 13 on average for the other approaches in the mystery genre and 12 for fairy-tales.

The summary of the results of the full game study is shown in Table~\ref{table:gameresults}.
As the comparisons made in this study are all made pairwise between our neural model and one of the baselines---they are presented in terms of what percentage of participants prefer the baseline game over the neural game.
Once again, as this is highly interrelated ranked data, we perform the Mann-Whitney U test between each of the pairs to ensure that the rankings differed significantly.
This is also indicated on the table.

In the mystery genre, the neural approach is generally preferred by a greater percentage of participants than the rules or random.
The human-made game outperforms them all.
A significant exception to is that participants thought that the rules-based game was more interesting than the neural game.
The trends in the fairy-tale genre are in general similar with a few notable deviations.
The first deviation is that the rules-based and random approaches perform significantly worse than neural in this genre.
We see also that the neural game is as coherent as the human-made game.

As in the previous study, we hypothesize that this is likely due to the rules-based approach being more suited to the mystery genre, which is often more mundane and contains less fantastical elements.
By extension, we can say that thematic commonsense in fairy-tales has less overlap with everyday commonsense than for mundane mysteries.
This has a few implications, one of which is that this theme specific information is unlikely to have been seen by OpenIE5 before.
This is indicated in the relatively improved performance of the rules-based model in this genre across in terms of both interestingness and coherence.
The genre difference can also be observed in terms of the performance of the random model.
This model also lacking when compared to our neural model across all the questions asked  especially in the fairy-tale setting.
This appears to imply that filling in gaps in the knowledge graph using thematically relevant information such as with AskBERT results in more interesting and coherent descriptions and games especially in settings where the thematic commonsense diverges from everyday commonsense.

\begin{table}[t]
\scriptsize
\centering
\begin{tabular}{l|l|l|l|l}
      Genre                      &      Questions           & Random & Rules & Human \\ \hline
\multirow{3}{*}{Mystery}    & Interesting     & 45     & 72*   & 69*   \\
                            & Coherence       & 36*    & 45*   & 69*   \\
                            & Resembles Genre & 45     & 38*   & 75*   \\ \hline
\multirow{3}{*}{Fairy-tale} & Interesting     & 42     & 37*   & 64*   \\ 
                            & Coherence       & 25*    & 25*   & 45    \\
                            & Resembles Genre & 25*    & 37*   & 69*    
\end{tabular}
\caption{Results of the full game evaluation participant study. *Indicates statistical significance ($p<0.05$).}
\label{table:gameresults}
\end{table}

\section{Conclusion}
Procedural world generation systems are required to be semantically consistent, comply with thematic and everyday commonsense understanding, and maintain overall interestingness.
We describe an approach that transform a linear reading experience in the form of a story plot into a interactive narrative experience.
Our method, AskBERT, extracts and fills in a knowledge graph using thematic commonsense and then uses it as a skeleton to flesh out the rest of the world.
A key insight from our human participant study reveals that the ability to construct a thematically consistent knowledge graph is critical to overall perceptions of coherence and interestingness particularly when the theme diverges from everyday commonsense understanding.

\newpage
\bibliographystyle{named}
\bibliography{ijcai20}
\end{document}